\pdfoutput=1
\documentclass[11pt]{article}
\usepackage[final]{acl}
\usepackage{times}
\usepackage{latexsym}
\usepackage[T1]{fontenc}
\usepackage[utf8]{inputenc}
\usepackage{microtype}
\usepackage{inconsolata}

\usepackage{graphicx}

\usepackage{booktabs} %
\usepackage{amsmath} %
\usepackage{utfsym} %
\usepackage{fontawesome5} %
\usepackage{footmisc} %
\usepackage{multirow} %
\usepackage{rotating} %

\definecolor{markerGreen}{RGB}{175,233,175} %
\definecolor{markerRed}{RGB}{255,170,170} %

\title{Detecting Referring Expressions in Visually Grounded Dialogue with Autoregressive Language Models}

\author{Bram Willemsen \and Gabriel Skantze \\
        Division of Speech, Music and Hearing \\
        KTH Royal Institute of Technology \\
        Stockholm, Sweden \\
        \texttt{\{bramw,skantze\}@kth.se}}

\begin{document}
\maketitle
\begin{abstract}
In this paper, we explore the use of a text-only, autoregressive language modeling approach for the extraction of referring expressions from visually grounded dialogue.
More specifically, the aim is to investigate the extent to which the linguistic context alone can inform the detection of mentions that have a (visually perceivable) referent in the visual context of the conversation.
To this end, we adapt a pretrained large language model (LLM) to perform a relatively course-grained annotation of mention spans in unfolding conversations by demarcating mention span boundaries in text via next-token prediction.
Our findings indicate that even when using a moderately sized LLM, relatively small datasets, and parameter-efficient fine-tuning, a text-only approach can be effective, highlighting the relative importance of the linguistic context for this task.
Nevertheless, we argue that the task represents an inherently multimodal problem and discuss limitations fundamental to unimodal approaches.

\end{abstract}

\section{Introduction}
In conversation, speakers often make reference to objects, events, or concepts. 
Words and phrases that are used with referential intent are known as \textit{referring expressions} (REs).
Effective communication hinges on the ability of the participants in the conversation to recognize these expressions and to determine what it is that they refer to, i.e., their \textit{referents}.
Within the context of a discourse, identification of an intended referent for a given RE may necessitate coreference resolution, i.e., the process of linking expressions that have the same referent.
To illustrate this need, consider the following hypothetical exchange, with coreferring expressions underlined:
\begin{quote}
($1$) \textbf{A}: Have you seen \underline{Schrödinger's cat}? \\
($2$) \textbf{B}: Yeah, here \underline{it} is. \\
($3$) \textbf{A}: \underline{It} is looking a bit worse for wear.
\end{quote}
Without access to the discourse context, ``\textit{it}'' and ``\textit{It}'' have indeterminate referents.
By having knowledge of the prior contributions to the conversation, it is clear that both pronouns are anaphors with ``\textit{Schrödinger's cat}'' as their antecedent.

\begin{figure}[t]
    \centering
    \includegraphics[width=1\linewidth]{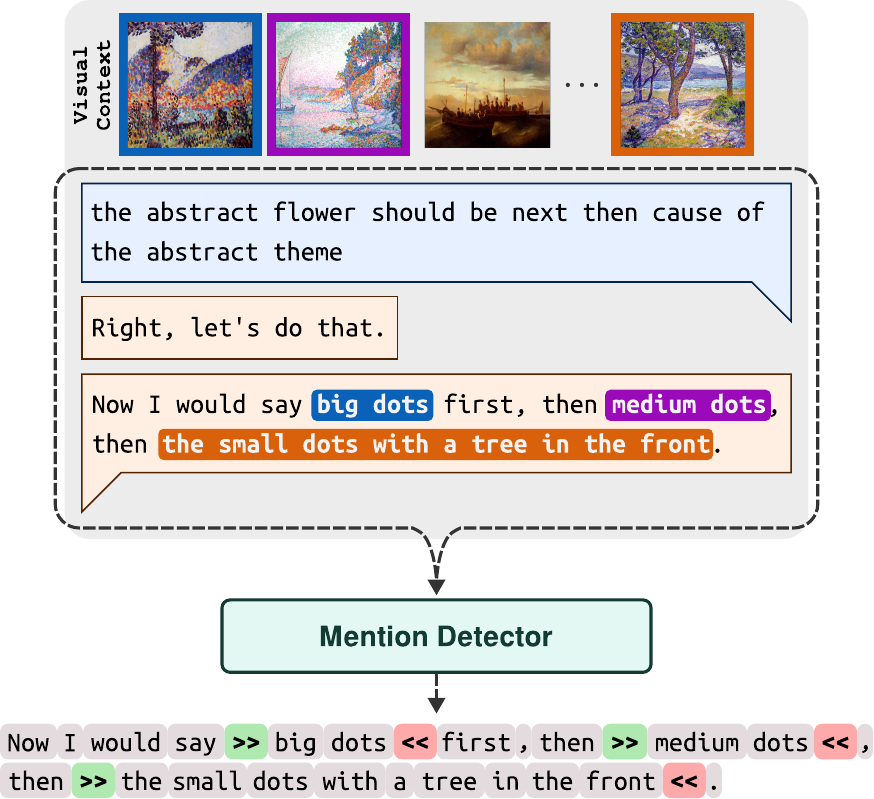}
    \caption{Visualization of the proposed mention detection method. 
    The \textbf{Mention Detector} takes as input the most recent dialogue message, preceded by the available dialogue history, and autoregressively outputs an annotated reproduction of the last message while inserting mention span boundary tokens (the start and end of mention spans are represented by \colorbox{markerGreen}{\textbf{\text{>}\text{>}}} and \colorbox{markerRed}{\textbf{\text{<}\text{<}}}, respectively) where appropriate.
    Excerpt shown is from a dialogue collected by \citet{willemsen_collecting_2022}. 
    Highlighted mentions in original dialogue and visual context with highlighted referent images are shown solely for illustrative purposes: the former is not available to the model at inference time, the latter neither at inference nor at training time.}
    \label{fig:mention-detection-diagram}
\end{figure}

The identification of REs, or \textit{mentions}\footnote{We use \textit{referring expression} and \textit{mention} interchangeably throughout this paper.}, in various types of discourse is a long-standing natural language processing (NLP) task commonly referred to as \textit{mention detection} (MD).
Simply put, when a given discourse is represented as a text document, the goal of MD is to identify any and all spans of text that refer to some predetermined type of referent, such as named entities or events. 

In this paper, we explore the problem of MD for conversation, specifically with a focus on the downstream purpose of reference resolution in visually grounded dialogue.
That is, our aim is to identify the REs that have a (visually perceivable) referent in the visual context of the conversation.
Of particular interest is the extent to which the linguistic context alone is able to inform predictions for what is arguably, inherently, a multimodal problem.
In addition, we experiment with different context windows to investigate how dialogue history affects MD performance.
The expectation is that providing access to additional linguistic context in the form of preceding messages will generally lead to increased performance.
To illustrate by example, whether the use of ``\textit{that}'' in the exclaimed utterance ``\textit{Take that!}'' is referential or instead part of a non-referential interjection cannot be known without additional context.

In line with recent work on generative information extraction \citep[see e.g.,][]{zhang_survey_2025}, we frame MD in visually grounded dialogue as an autoregressive language modeling problem.
More specifically, we propose to train a model to generate annotated reproductions of utterances: for a given utterance, in the process of generating a copy of the original message content, the model is expected to insert span boundary tokens indicating the start and end of mention spans, when and where appropriate.
An illustration of the proposed approach is shown in Figure \ref{fig:mention-detection-diagram}.
Our experiments involve the parameter-efficient fine-tuning \citep{dettmers_qlora_2023} of a large language model (LLM) on annotated conversations from two different visually grounded dialogue datasets, namely \textsc{A Game Of Sorts} \citep[\textsc{AGOS},][]{willemsen_collecting_2022} and \textsc{PhotoBook} \citep[\textsc{PB},][]{haber_photobook_2019}. 
For \textsc{AGOS}, we make use of the mention annotations from \citet{willemsen_resolving_2023}.
For \textsc{PB}, we adopt a similar annotation protocol to manually create the required mention annotations for a subset of the dataset.\footnote{\url{https://github.com/willemsenbram/mention-detection-vgd}, \href{https://doi.org/10.5281/zenodo.15500581}{doi:10.5281/zenodo.15500581} \label{fn:repo}}

Results of our experiments with the \textsc{8B}-parameter variant of \textsc{Llama 3.1} \citep{grattafiori_llama_2024} are promising, suggesting that the linguistic context can be relatively revealing for our purpose.
Note that our findings are in spite of the fact that our datasets are relatively small, our LLM is relatively moderately sized, and our fine-tuning is parameter-efficient.
Nevertheless, we must contend with some limitations that are fundamental to unimodal approaches to multimodal problems, as well as the nature of the referential language in task-oriented dialogues. 
We provide additional discussion on these matters.

\section{Background}
MD has long been an essential component, or the central focus, of systems addressing various NLP tasks, such as named entity recognition \citep[e.g.,][]{lample_neural_2016,devlin_bert_2019,strakova_neural_2019}, event detection \citep[e.g.,][]{lai_event_2020}, and coreference resolution \citep[e.g.,][]{lee_deterministic_2013,poesio_anaphora_2018}.
Earlier, rule-based approaches to MD were frequently built atop a dependency parse of a text, and would, over time, incorporate increasingly more powerful statistical models into the pipeline \citep[e.g.,][]{florian_improving_2010,lee_deterministic_2013}.
The required sophistication of the approach generally depended on the downstream task.
For coreference resolution, for example, simple heuristics leading to high recall would suffice if other parts of the system could compensate with higher precision \citep[e.g.,][]{lee_deterministic_2013}.
Interestingly, comparisons between different coreference resolution systems have often been conducted on the basis of gold, instead of predicted, mentions. 
This effectively side-steps MD in an effort to focus on isolating the system's downstream performance. 
However, there tend to be notable performance gaps between these idealized and the realistic scenarios.
As \citet{poesio_computational_2023} note, generally, the overall performance of a coreference resolution system has been contingent on the accuracy of the output from its MD component. 

Following advances in neural language modeling, approaches to MD based on neural models \citep[e.g.,][]{lample_neural_2016,poesio_anaphora_2018,devlin_bert_2019,strakova_neural_2019,lai_event_2020,yu_neural_2020} have gradually superseded the earlier methods.
These increasingly more data-driven methods promised to do away with the need for extensive feature engineering.
Particularly consequential has been the adoption of general purpose, pretrained language models based on the Transformer architecture \citep{vaswani_attention_2017}, examples of which include the encoder-only BERT \cite{devlin_bert_2019} and the decoder-only GPT \citep{radford_improving_2018}.
BERT-based representations have been the backbone of numerous NLP systems, including those that deal with MD \citep[e.g.,][]{devlin_bert_2019,strakova_neural_2019,yu_neural_2020}.

Of particular interest here are the autoregressive LLMs at the heart of most work on generative information extraction \cite[see e.g.,][]{zhang_survey_2025}. 
Various studies have shown that framing tasks involving structured predictions as autoregressive language modeling problems can be effective \citep[e.g.,][]{cao_autoregressive_2021,liu_autoregressive_2022,deuser_informed_2024}.
Given an unstructured text, the model is trained to return, via next-token prediction, a structured representation of the input. 
Although the feasibility of this approach has been shown for commonly used benchmarks that involve some form of MD \citep[e.g.,][]{kim_genia_2003-1,tjong_kim_sang_introduction_2003}, to the best of our knowledge, it has yet to be applied to visually grounded dialogue.
In this paper, we explore to what extent we can adapt a pretrained LLM via parameter-efficient fine-tuning \citep{hu_lora_2022,dettmers_qlora_2023} to the task of MD in visually grounded dialogue using this approach.

\section{Method}

\subsection{Problem description}
In general, the goal of MD is to identify all expressions in a document $D$ that satisfy some prescribed definition of a \textit{mention}.
When $D$ is a \textit{visually grounded} dialogue, we define it as $D = (V, L)$, where $V$ is the visual context and $L$ the linguistic context of the conversation.
A dialogue is considered visually grounded when $L$ contains one or more references to $V$.
That is, within the linguistic context, there exists one or more expressions that have a (visually perceivable) referent that is present in the visual context of the conversation.

\subsection{Task definition}
In this work, we consider MD in visually grounded dialogue to be the task of identifying all expressions in $L$ for which there exists a referent in $V$.
Here, we focus on visually grounded dialogues of which $V$ is composed of a set of $v$ independent images, $V = \{I_{1}, I_{2}, ..., I_{v}\}$.
The linguistic context $L$ can be represented as a sequence of $n$ utterances\footnote{We use \textit{utterance} and \textit{message} interchangeably throughout this paper.}, $L = (u_{1}, u_{2}, ..., u_{n})$. 
In turn, each utterance $u_{i}$ can be represented as a sequence of $m_i$ tokens, $u_{i} = (t_{i1}, t_{i2}, ..., t_{im_{i}})$.
We think of mentions in terms of spans.
We can define a mention span as a contiguous subsequence of tokens from an utterance $u_{i}$, denoted as $(t_{ij}, ..., t_{ik}) \subseteq u_i$, where $1 \leq j \leq k \leq m_i$.  
Together, these tokens constitute an expression that (indirectly) refers to one or more of the images.
Note that in contrast with other types of documents, dialogue is interactive and contributions to $L$ are cumulative, happening over time.
It is important to account for the incremental nature of conversation when addressing this task.

\subsection{Proposed approach}
Core to our approach is the framing of MD in visually grounded dialogue as a next-token prediction task.
Given the incremental nature of conversation, we process each dialogue at the utterance level, prepending to each utterance a token indicating the speaker.
For a given utterance $u_{i}$, we train an autoregressive language model $f$ to reproduce exactly the original content of $u_{i}$, but with span boundary tokens inserted if and where appropriate to indicate the start and end of mention spans.

Crucially, however, we propose to condition the generation of the target sequence ${u_{i}}'$ not only on the current utterance $u_{i}$, but also on additional preceding linguistic context, i.e., the available dialogue history, as prior messages may inform predictions.
When considering prior messages in the modeling process, we can define the generation of ${u_{i}}'$ as ${u_{i}}' = f(u_{i}, H)$, where $H$ is the dialogue history available to the model.\footnote{We must note that for the experiments reported in this paper, we found that repeating utterance ${u_{i}}$ in the input to the model had a positive impact on downstream performance; a slight deviation from the more general definition provided here. For an example of the formatting of training samples, see Appendix \ref{sec:training-example}.}
The available dialogue history $H$ is defined as a contiguous subsequence of utterances from $L$, denoted as $H = ( u_{i-h}, u_{i-h+1} ..., u_{i-1} )$, where $0 \leq h \leq w$, where $h$ is the number of prior messages available to the model and $w$ is an optionally predefined maximum number of preceding messages to be considered. 
For a visualization of the proposed approach, see Figure \ref{fig:mention-detection-diagram}.

\section{Experiments}
The language modeling experiments presented in this paper involve the fine-tuning of pretrained models on dialogues from two different, though closely related, visually grounded dialogue tasks, namely \textsc{A Game Of Sorts} \citep[\textsc{AGOS},][]{willemsen_collecting_2022} and \textsc{PhotoBook} \citep[\textsc{PB},][]{haber_photobook_2019}.
We first perform cross-validation to score MD performance on each dataset separately.
We then assess cross-dataset transfer by training on one dataset and testing on the other.
In addition, we investigate the effects of dialogue history on MD performance, i.e., whether the model benefits from having access to preceding messages when making its predictions, by experimenting with different context window sizes, i.e., providing access to different numbers of preceding messages.
Finally, as points of comparison, we assess the MD performance of a baseline based on noun phrase (NP) extraction using constituency parsing, as well as that of an encoder-only LLM fine-tuned for sequence labeling.

\subsection{Data}
Both \textsc{AGOS} and \textsc{PB} are tasks designed around eliciting repeated references to various sets of real-world images---such as those found in the MS COCO \citep{lin_microsoft_2014} and Open Images \citep{kuznetsova_open_2020} datasets---in conversational settings.
Moreover, both tasks have a deliberate asymmetry in their visual contexts that participants have to overcome to successfully complete the task.
This ensured that speakers would produce non-trivial REs that made reference to the images' visual content.

\begingroup
    \renewcommand{\arraystretch}{1}
    \begin{table}[t!]
    \centering
    \footnotesize
        \begin{tabular}{l|r|r}\toprule
            & \textbf{\textsc{AGOS}} & \textbf{\textsc{PB-GOLD}} \\ 
            \midrule
            $\#$ Dialogues & $15$ & $50$ \\
            $\#$ Messages (\faComment[regular]) & $1,800$ & $3,335$ \\
            $\#$ Mentions (\faHighlighter) & $1,486$ & $2,111$ \\
            $\#$ Characters (\text{\faFont}) & $86,516$ & $96,774$ \\
            $\#$ Words (\text{\faFont}\text{\faBold}) & $19,843$ & $22,889$ \\
            \midrule
            $\%$ \faComment[regular] with \faHighlighter & $60.33\%$ & $61.02\%$ \\
            $\%$ \faComment[regular] with > $1$ \faHighlighter & $17.94\%$ & $1.95\%$ \\
            \midrule
            $\#$ \text{\faFont} in \faHighlighter & $27,574$ & $61,771$ \\
            $\%$ \text{\faFont} in \faHighlighter $:$ \text{\faFont} in \faComment[regular] & $31.87\%$ & $63.83\%$ \\
            \midrule
            $\#$ \text{\faFont}\text{\faBold} in \faHighlighter & $5,708$ & $12,880$ \\
            $\%$ \text{\faFont}\text{\faBold} in \faHighlighter $:$ \text{\faFont}\text{\faBold} in \faComment[regular] & $28.77\%$ & $56.27\%$ \\
            \midrule
            $\bar{X}$ \text{\faFont} in \faComment[regular] & $48.06$ {\scriptsize($43.57$)} & $29.02$ {\scriptsize($24.83$)} \\
            $\bar{X}$ \text{\faFont} in \faHighlighter & $18.56$ {\scriptsize($15.76$)} & $29.26$ {\scriptsize($23.35$)}  \\
            \midrule
            $\bar{X}$ \text{\faFont}\text{\faBold} in \faComment[regular] & $11.02$ {\scriptsize($9.52$)} & $6.86$ {\scriptsize($5.40$)} \\
            $\bar{X}$ \text{\faFont}\text{\faBold} in \faHighlighter & $3.84$ {\scriptsize($3.20$)} & $6.10$ {\scriptsize($4.86$)} \\
            \bottomrule
        \end{tabular}
        \caption{\label{table:descriptives-datasets}
            Descriptive statistics for the \textsc{AGOS} and \textsc{PB-GOLD} datasets.
            \textit{Note}. Explanation of symbols and abbreviations: \text{\faComment[regular]} = Messages; \text{\faHighlighter} = Mentions; \text{\faFont} = Characters; \text{\faFont}\text{\faBold} = Words; $\bar{X}$ = average (mean).
            Standard deviation between brackets.
            Scores and standard deviations are rounded to the nearest hundredth.}
    \end{table}
\endgroup

\begingroup
\renewcommand{\arraystretch}{1}
    \begin{table*}[t!]
    \centering
    \footnotesize
        \begin{tabular}{l|l|cccc|cccc}\toprule
            & & \multicolumn{4}{c}{\textsc{\textbf{AGOS}}} & \multicolumn{4}{|c}{\textsc{\textbf{PB-GOLD}}} 
            \\\cmidrule(lr){3-6}\cmidrule(lr){7-10}
            & & $\mathbf{0}$ & $\mathbf{3}$ & $\mathbf{7}$ & $\mathbf{19}$ & $\mathbf{0}$ & $\mathbf{3}$ & $\mathbf{7}$ & $\mathbf{19}$ \\\midrule
            \parbox[t]{2mm}{\multirow{4}{*}{\rotatebox[origin=c]{90}{\textsc{Llama}}}} & P & $.896$ {\scriptsize($.03$)} & $.922$ {\scriptsize($.02$)} & $.919$ {\scriptsize($.02$)} & \underline{$.923$} {\scriptsize($.03$)} & $.933$ {\scriptsize($.02$)} & $.936$ {\scriptsize($.03$)} & $.940$ {\scriptsize($.02$)} & \underline{$.943$} {\scriptsize($.02$)} \\
            & R & $.835$ {\scriptsize($.04$)} & $.865$ {\scriptsize($.03$)} & $.883$ {\scriptsize($.03$)} & \underline{$.884$} {\scriptsize($.03$)} & $.927$ {\scriptsize($.01$)} & $.925$ {\scriptsize($.01$)} & $.934$ {\scriptsize($.01$)} & \underline{$.937$} {\scriptsize($.02$)} \\
            & \textit{F}$_{1}$ & $.863$ {\scriptsize($.02$)} & $.892$ {\scriptsize($.01$)} & $.900$ {\scriptsize($.01$)} & \underline{$.902$} {\scriptsize($.01$)} & $.930$ {\scriptsize($.01$)} & $.930$ {\scriptsize($.02$)} & $.937$ {\scriptsize($.02$)} & \underline{$.940$} {\scriptsize($.02$)} \\
            & $J$ & $.811$ {\scriptsize($.03$)} & $.849$ {\scriptsize($.03$)} & $.856$ {\scriptsize($.02$)} & \underline{$.858$} {\scriptsize($.02$)} & $.921$ {\scriptsize($.01$)} & $.922$ {\scriptsize($.01$)} & \underline{$.933$} {\scriptsize($.01$)} & \underline{$.933$} {\scriptsize($.01$)} \\\midrule
            \midrule
            \parbox[t]{2mm}{\multirow{4}{*}{\rotatebox[origin=c]{90}{\textsc{M-BERT}}}} & P & $.827$ {\scriptsize($.04$)} & $.842$ {\scriptsize($.03$)} & $.843$ {\scriptsize($.03$)} & \underline{$.863$} {\scriptsize($.04$)} & $.916$ {\scriptsize($.02$)} & $.918$ {\scriptsize($.02$)} & $.924$ {\scriptsize($.01$)} & \underline{$.930$} {\scriptsize($.02$)} \\
            & R & $.812$ {\scriptsize($.05$)} & $.835$ {\scriptsize($.03$)} & $.837$ {\scriptsize($.04$)} & \underline{$.853$} {\scriptsize($.01$)} & $.909$ {\scriptsize($.01$)} & $.912$ {\scriptsize($.01$)} & $.908$ {\scriptsize($.01$)} & \underline{$.917$} {\scriptsize($.02$)} \\
            & \textit{F}$_{1}$ & $.819$ {\scriptsize($.04$)} & $.838$ {\scriptsize($.02$)} & $.839$ {\scriptsize($.02$)} & \underline{$.857$} {\scriptsize($.02$)} & $.912$ {\scriptsize($.02$)} & $.915$ {\scriptsize($.01$)} & $.916$ {\scriptsize($.01$)} & \underline{$.924$} {\scriptsize($.02$)} \\
            & $J$ & $.786$ {\scriptsize($.04$)} & $.815$ {\scriptsize($.02$)} & $.814$ {\scriptsize($.02$)} & \underline{$.825$} {\scriptsize($.01$)} & $.909$ {\scriptsize($.01$)} & $.914$ {\scriptsize($.01$)} & $.913$ {\scriptsize($.01$)} & \underline{$.920$} {\scriptsize($.01$)} \\            
            \bottomrule
        \end{tabular}
        \caption{\label{table:agos-photobook-cross-validation-results}
        Cross-validated mention detection performance of fine-tuned \textsc{Llama 3.1 8B} (\textsc{Llama}, top) and \textsc{ModernBERT}-large (\textsc{M-BERT}, bottom) on \textsc{AGOS} and \textsc{PB-GOLD} for four different context windows, i.e., $0$, $3$, $7$, and $19$ preceding messages.
        \textit{Note}. P = Precision; R = Recall; \textit{F}$_{1}$ = \textit{F}$_{1}$ score; $J$ = Jaccard index.
        Scores are rounded to the nearest thousand, standard deviations to the nearest hundredth.}
    \end{table*}
\endgroup

\subsubsection{A Game Of Sorts (\textsc{AGOS})}
\textsc{AGOS} is a collaborative image ranking task. 
Two participants are shown a set of nine images which they are asked to rank, in descending order and one at a time, based on a given sorting criterion.
The goal of the task is for the participants to, through conversation, arrive at a ranking which both deem satisfactory.
Although both participants see the same set of images, they cannot see each other's perspective. 
The position of the images on their respective screens has been randomized, forcing the participants to refer to the images by referencing their visual content.
To ensure repeated mentions of the same referents, the task is performed over multiple (four) rounds, and the same set of images is used each round.

The \textsc{AGOS} dataset consists of $15$ dialogues.
Each \textsc{AGOS} image set consists of nine images from the same of one of five image categories, namely cars, dogs, paintings, pastries, or phones.
Three dialogues were collected per image category.

\subsubsection{PhotoBook (\textsc{PB})}
\textsc{PB} is a collaborative image identification task. 
Two participants are shown partially dissimilar sets of six visually similar images; some of the images will be shown to both participants, while others are shown to only one of the participants.
Each participant has three of their six images highlighted.
The goal of the task is for the participants to, through conversation and without seeing each other's perspective, identify for these highlighted images whether or not they have them in common.
To ensure repeated mentions of the same referents, the task is performed over multiple (five) rounds, and while the set of images shown to participants changes from round to round, the image sets are constructed in such a way that each image is shown multiple times to at least one of the participants.

The \textsc{PB} dataset consists of $2.5$K dialogues.
Each \textsc{PB} image set, as shown to each participant, consists of six images that prominently feature two objects, each object belonging to a different image category.
These two image categories form the ``image domain'' of the conversation; each image shown throughout the interaction will feature at least one object from each category.
For our experiments, we make use of the so-called \textsc{PB-GOLD} subset, as referenced in \citet{takmaz_less_2022}, which consists of $50$ dialogues for which the authors have provided some annotations at the utterance level.

\subsubsection{Mention annotations}
In this work, we make use of the manually annotated mention spans from \citet{willemsen_resolving_2023}.
These spans indicate the linguistic expressions that have a (visually perceivable) referent in the visual context of the conversation.
More specifically, these are either singletons or REs that are part of an identity relation with other mentions in the linguistic context that have one or more of the images as their referents.
For the annotation of the mention spans, \citet{willemsen_resolving_2023} were aided by speakers' self-annotations, as participants were required to indicate whether or not a message was meant to include one or more references to one or more of the images.
In the messages which contained such references, the longest, most specific spans with images as their referents were marked.
The resulting annotations are relatively course-grained.
We adopt this protocol for our annotation of the \textsc{PB-GOLD} dialogues.
Although \textsc{PB} has no self-annotations, referential ambiguities can be resolved by scrutiny of the full dialogue context.
We report descriptive statistics of both datasets in Table \ref{table:descriptives-datasets}.

\subsection{Model specifications}
For each experiment involving the proposed autoregressive language modeling approach, we fine-tune \textsc{Llama 3.1 8B} \citep{grattafiori_llama_2024} using QLoRA \citep{dettmers_qlora_2023} on a single 24GB NVIDIA GeForce RTX 3090.
We calculate the loss only over tokens of the target message, masking the loss over tokens that are part of the preceding dialogue context. 
We make use of the model's existing vocabulary for any special tokens, such as those indicating span boundaries.
Fine-tuned model output is generated using constrained decoding.
That is, at every time step we dynamically restrict the vocabulary, where the allowed tokens include the next token from the input utterance and any valid special tokens. %
Hyperparameters are listed in Table \ref{table:hyperparameters-llama} in Appendix \ref{sec:hyperparameters}.
For an example of the formatting of training samples for fine-tuning, see Appendix \ref{sec:training-example}.
For additional implementation details, we refer the reader to our repository.\footref{fn:repo}

\subsubsection{Baselines}
\noindent\textbf{NP extraction using constituency parsing}
As mentions are predominantly NPs, we opt for a simple baseline model that automatically extracts NPs from the dialogues using the constituency parser from the Stanza toolkit \cite{qi_stanza_2020}.
The backbone of this parser is ELECTRA-large \citep{clark_electra_2020} trained on a revised version of the third release of the Penn Treebank \citep{marcus_building_1993}.
We extract the most expansive spans, but discard certain candidate phrases.
For instance, as the dialogues involve text-based conversations in which the participants are not able to see each other, we can disregard various personal pronouns (e.g., ``\textit{I}'', ``\textit{you}'', ``\textit{me}'') as these were not considered to be mentions here.

\noindent\textbf{Sequence labeling with \textsc{ModernBERT}}
It has been common practice to treat problems that center on the detection of spans in text, such as MD, as sequence labeling tasks \citep[e.g.,][]{lample_neural_2016}.
When given a sequence (of tokens), the objective is to assign each element a label such that span boundaries can be inferred.
Tag sets are frequently based on the IOB format \citep{ramshaw_text_1995}: the B tag indicates that an element begins a span, the I tag indicates that an element is inside of a span, and the O tag indicates that an element is outside of a span.
For our purpose, we adopt the IOB tag set and fine-tune \textsc{ModernBERT}-large \citep{warner_smarter_2024} to predict for each token of a given utterance the correct label.
As the name suggests, \textsc{ModernBERT} is a more recent encoder-only LLM that improves upon the original BERT architecture.
Similar to the \textsc{Llama}-based experiments, for the experiments that are meant to demonstrate the effects of dialogue history on downstream performance, we provide preceding messages as context, masking the loss over all labels except those of the target message.
Each model is fine-tuned on a single 24GB NVIDIA GeForce RTX 3090.
Hyperparameters are listed in Table \ref{table:hyperparameters-modernbert} in Appendix \ref{sec:hyperparameters}.
For additional implementation details, we refer the reader to our repository.\footref{fn:repo}

Note that in this formulation of the problem using the basic IOB format, it is not possible to accurately label nested mentions.
However, there are very few cases of nesting in the datasets used for the experiments reported in this paper.
Therefore, this shortcoming has negligible impact on the current evaluation of the approach.

\subsection{Evaluation}
Our first experiments involve cross-validation on both datasets.
We evaluate using the same five-fold cross-validation protocol adopted by prior work on the \textsc{AGOS} dataset \citep{willemsen_resolving_2023,willemsen_referring_2024}, which partitions the dataset along its five image sets.
We similarly perform five-fold cross-validation on the \textsc{PB-GOLD} dataset.
However, as there is no predefined, deterministic split for \textsc{PB-GOLD}, we split the data randomly.
Our second set of experiments concerns an investigation into cross-dataset transfer.
This means that we fine-tune models on the entirety of \textsc{AGOS} and test on the entirety of \textsc{PB-GOLD}, and vice versa.

In addition, we test the effects of dialogue history on MD performance.
For each of the aforementioned experiments, we fine-tune models for four different context windows, $0$, $3$, $7$, and $19$, meaning the models have access to no, three, seven, or 19 preceding messages, respectively.

\subsubsection{Metrics}
We measure mention detection performance in terms of precision, recall, \textit{F}$_{1}$ score, and intersection over union of ground truth (gold spans) and predicted mention spans at the character level (i.e., Jaccard index).\footnote{Character-level evaluation avoids tokenization issues when span boundary tokens are placed within words.} %

We calculate precision, recall, and \textit{F}$_{1}$ scores based on exact mention span matches.
This means that a predicted mention is considered a true positive only if it matches a gold span exactly and is treated as a false positive otherwise.
Conversely, a ground truth mention for which there is no exact matching prediction is considered a false negative.

We use a measure based on the Jaccard index to score the extent to which ground truth and predicted mention spans overlap, which permits the scoring of partial matches.
For each message, we find the optimal assignment of predicted and ground truth spans based on the number of corresponding character indices. 
We calculate the Jaccard index for each pair of matched spans. 
In the event that no match exists---that is, there is no overlap between a ground truth mention and any of the predicted spans (false negative), or there exists no ground truth mention for a predicted span (false positive)---, the score for this particular span is $0$.

All the aforementioned mention detection metrics are bound $[0, 1]$, with higher scores indicating better performance.

\begin{figure}[t!]
    \centering
    \includegraphics[trim={0.35cm 0.4cm -0.7cm 0.3cm},clip,width=1\linewidth]{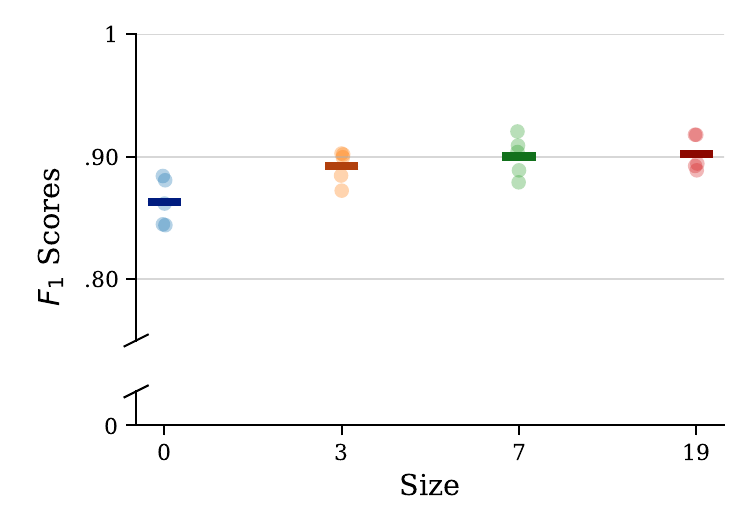}
    \caption{Mention detection performance of fine-tuned \textsc{Llama 3.1 8B} in terms of $F_{1}$ scores $[0, 1]$ as a function of the size of the context window, i.e., the maximum number of preceding messages considered from the available dialogue history. 
    Shown are results of each fold (dots) and their average (bar) for four different context windows, i.e., $0$, $3$, $7$, and $19$.} %
    \label{fig:f1-scores-plot}
\end{figure}

\section{Results}
Before reporting the results of our fine-tuning experiments, we first highlight some of the descriptive statistics reported in Table \ref{table:descriptives-datasets} to aid in understanding the composition of the data.
As shown in Table \ref{table:descriptives-datasets}, \textsc{PB-GOLD} contains over three times more dialogues than \textsc{AGOS}.
However, on average, the \textsc{AGOS} dialogues are considerably longer and have almost twice as many messages per dialogue.
While the percentage of messages with mentions is comparable, \textsc{AGOS} has a much higher rate of messages that contain more than one mention than \textsc{PB-GOLD}.
Nevertheless, mentions make up notably less of the overall content of the \textsc{AGOS} dialogues than of the \textsc{PB-GOLD} dialogues; the number of characters and words dedicated to mentions relative to the total number of characters and words in the messages is substantially lower for \textsc{AGOS} than for \textsc{PB-GOLD}.
Finally, the average \textsc{AGOS} mention is shorter than the average \textsc{PB-GOLD} mention.

\begingroup
    \renewcommand{\arraystretch}{1}
    \begin{table}[t!]
    \centering
    \footnotesize
        \begin{tabular}{l|ccccc|c}\toprule
            & \faCarSide & \faDog & \faPaintBrush & \faBirthdayCake & \faMobile* & $\bar{X}$
            \\\midrule
            P & $.881$ & $.954$ & $.934$ & $.912$ & $.933$ & $.923$ {\scriptsize($.03$)} \\
            R & $.897$ & $.842$ & $.902$ & $.924$ & $.855$ & $.884$ {\scriptsize($.03$)} \\
            \textit{F}$_{1}$ & $.889$ & $.894$ & $.918$ & $.918$ & $.892$ & $.902$ {\scriptsize($.01$)} \\
            $J$ & $.853$ & $.823$ & $.881$ & $.874$ & $.857$ & $.858$ {\scriptsize($.02$)} \\
            \bottomrule
        \end{tabular}
        \caption{\label{table:results-19-all}
            Cross-validated mention detection performance of fine-tuned \textsc{Llama 3.1 8B} on \textsc{AGOS} based on a context window of $19$, i.e., a dialogue history consisting of $19$ preceding messages.
            Results are shown for each fold as well as their average ($\bar{X}$).
            \textit{Note}. P = Precision; R = Recall; \textit{F}$_{1}$ = \textit{F}$_{1}$ score; $J$ = Jaccard index; Symbols represent folds: $\text{\faCarSide}$ = Cars; $\text{\faDog}$ = Dogs; $\text{\faPaintBrush}$ = Paintings; $\text{\faBirthdayCake}$ = Pastries; $\text{\faMobile*}$ = Phones.
            Standard deviation between brackets.
            Scores are rounded to the nearest thousand, standard deviations to the nearest hundredth.}
    \end{table}
\endgroup

\subsection{Cross-validation}
Shown in Table \ref{table:agos-photobook-cross-validation-results} are the cross-validated results from fine-tuning and evaluating models on the \textsc{AGOS} and \textsc{PB-GOLD} datasets.
For each context window, scores are reported as averages over all folds for each MD performance metric.
In addition, the results reported in Table \ref{table:results-19-all} are from fine-tuning and evaluating \textsc{Llama} on the \textsc{AGOS} dataset using the maximum context window size we considered for this work, i.e., a context window of size $19$.
In Table \ref{table:results-19-all}, scores are shown per fold in addition to their averages over all folds, for each MD performance metric.
We found that, despite some variance between folds, scores resulting from fine-tuning \textsc{Llama} were relatively high overall.
In comparison, the performance of \textsc{ModernBERT} is relatively competitive, but it does lag behind that of \textsc{Llama}.
The observed results suggest that the models were, on average, somewhat more performant on the \textsc{PB-GOLD} than they were on the \textsc{AGOS} data.
Moreover, we observed that the models generally benefited from an increase in context window size; on average, we found that providing the models with a greater number of preceding messages increased MD performance, but noted that there were diminishing returns.
The observed trend was somewhat more apparent for \textsc{AGOS} than for \textsc{PB-GOLD}.
Figure \ref{fig:f1-scores-plot} provides a visualization of this trend based on the $F_{1}$ scores for \textsc{AGOS}.

\begingroup
\renewcommand{\arraystretch}{1}
    \begin{table*}[t!]
    \centering
    \footnotesize
        \begin{tabular}{l|l|cccc|cccc}\toprule
            & & \multicolumn{4}{c}{\textsc{\textbf{AGOS} $\rightarrow$ \textbf{PB-GOLD}}} & \multicolumn{4}{|c}{\textsc{\textbf{PB-GOLD} $\rightarrow$ \textbf{AGOS}}} 
            \\\cmidrule(lr){3-6}\cmidrule(lr){7-10}
            & & $\mathbf{0}$ & $\mathbf{3}$ & $\mathbf{7}$ & $\mathbf{19}$ & $\mathbf{0}$ & $\mathbf{3}$ & $\mathbf{7}$ & $\mathbf{19}$ \\\midrule
            \parbox[t]{2mm}{\multirow{4}{*}{\rotatebox[origin=c]{90}{\textsc{Llama}}}} & P & $.798$ & $.859$ & $.864$ & \underline{$.886$} & $.775$ & $.803$ & $.810$ & \underline{$.820$} \\
            & R & $.806$ & $.838$ & \underline{$.858$} & $.845$ & $.687$ & $.676$ & $.713$ & \underline{$.744$} \\
            & \textit{F}$_{1}$ & $.802$ & $.848$ & $.861$ & \underline{$.865$} & $.728$ & $.734$ & $.758$ & \underline{$.780$} \\
            & $J$ & $.777$ & $.816$ & $.834$ & \underline{$.839$} & $.668$ & $.666$ & $.694$ & \underline{$.722$} \\\midrule
            \midrule
            \parbox[t]{2mm}{\multirow{4}{*}{\rotatebox[origin=c]{90}{\textsc{M-BERT}}}} & P & $.725$ & $.768$ & $.778$ & \underline{$.795$} & $.707$ & $.735$ & $.774$ & \underline{$.777$} \\
            & R & $.650$ & $.694$ & $.687$ & \underline{$.735$} & $.610$ & $.641$ & $.704$ & \underline{$.723$} \\
            & \textit{F}$_{1}$ & $.685$ & $.729$ & $.730$ & \underline{$.764$} & $.655$ & $.685$ & $.737$ & \underline{$.749$} \\
            & $J$ & $.662$ & $.704$ & $.698$ & \underline{$.737$} & $.595$ & $.616$ & $.665$ & \underline{$.688$} \\            
            \bottomrule
        \end{tabular}
        \caption{\label{table:agos-photobook-transfer-results}
        Mention detection performance of fine-tuned \textsc{Llama 3.1 8B} (\textsc{Llama}, top) and \textsc{ModernBERT}-large (\textsc{M-BERT}, bottom) in cross-data transfer experiments for four different context windows, i.e., $0$, $3$, $7$, and $19$ preceding messages.
        \textsc{\textbf{AGOS} $\rightarrow$ \textbf{PB-GOLD}} indicates training on \textsc{AGOS} and testing on \textsc{PB-GOLD}; \textsc{\textbf{PB-GOLD} $\rightarrow$ \textbf{AGOS}} indicates training on \textsc{PB-GOLD} and testing on \textsc{AGOS}.
        \textit{Note}. P = Precision; R = Recall; \textit{F}$_{1}$ = \textit{F}$_{1}$ score; $J$ = Jaccard index.
        Scores are rounded to the nearest thousand, standard deviations to the nearest hundredth.}
    \end{table*}
\endgroup

\subsection{Cross-dataset transfer}
Table \ref{table:agos-photobook-transfer-results} shows results from fine-tuning models on \textsc{AGOS} and testing on \textsc{PB-GOLD} (\textsc{AGOS $\rightarrow$ PB-GOLD}), and vice versa (\textsc{PB-GOLD $\rightarrow$ AGOS}).
Although scores were shown to trail those of the cross-validation experiments, the observed MD performance was still indicative of a relatively high degree of successful transfer overall.
Again, \textsc{Llama}'s performance was shown to exceed that of \textsc{ModernBERT}.
A noteworthy observation was that \textsc{AGOS $\rightarrow$ PB-GOLD} consistently resulted in higher scores than \textsc{PB-GOLD $\rightarrow$ AGOS} on all MD performance metrics.
Similarly to results from our cross-validation experiments, we observed that, on average, an increase in the size of the context window tended to result in improved performance.
These findings suggest that providing the models with at least some preceding messages can already be beneficial.

\subsection{Comparison with NP extraction}
The results reported in Table \ref{table:results-stanza-baseline} show the MD performance of a method based on constituency parsing for the automatic extraction of NPs.
Although recall may seem relatively high considering that the focus of this baseline model was solely on NP extraction, it bears repeating that most mentions tend to be NPs, though they are not always presented in a straightforward, parsable manner or context. 
Perhaps unsurprisingly, especially when comparing against our proposed approach, this naive method for MD is relatively imprecise, as the false positive rate ends up being relatively high when predicting virtually all NPs to be referential in nature.

\subsection{Error analysis}
When examining the output generated by \textsc{Llama}, we found various errors to be consistent between the different context windows.
Although the models appeared to be relatively robust against the noise in the input, certain mentions were partially, or entirely, missed, as a result of ungrammatical phrasing.
For partial matches, we observed some recurring errors in relation to structural ambiguities, leading to the exclusion of relative clauses or prepositional phrases, and the splitting of single into multiple mentions or the merging of multiple mentions into a single span.
Furthermore, we found instances of ambiguous pronoun usage to be relatively frequent among errors, such as in the phrases ``\textit{let's go for it}'' and ``\textit{let's do it}'', in which the use of ``\textit{it}'' is referential, but it is not recognized as such without additional context.
Interestingly, providing access to preceding messages ends up resolving the inaccuracy for the former and not for the latter, even though these seem to be very similar cases on the surface.
Conversely, we also observed cases where usage of (pro)nouns was incorrectly predicted to be referential.
Again, some of these errors were resolved by providing the model access to the dialogue history.

\begingroup
    \renewcommand{\arraystretch}{1}
    \begin{table}[t!]
    \centering
    \footnotesize
        \begin{tabular}{l|c|c}\toprule
            & \textsc{\textbf{AGOS}} & \textsc{\textbf{PB-GOLD}} 
            \\\midrule
            P & $.411$ & $.377$ \\
            R & $.764$ & $.607$ \\
            \textit{F}$_{1}$ & $.535$ & $.465$ \\
            $J$ & $.453$ & $.530$ \\
            \bottomrule
        \end{tabular}
        \caption{\label{table:results-stanza-baseline}
            Mention detection performance of the Stanza NP extraction baseline.
            \textit{Note}. P = Precision; R = Recall; \textit{F}$_{1}$ = \textit{F}$_{1}$ score; $J$ = Jaccard index. 
            Standard deviation between brackets. 
            Scores are rounded to the nearest thousand, standard deviations to the nearest hundredth.}
    \end{table}
\endgroup

\section{Discussion}
In this paper, we explored the potential of an approach to mention detection (MD) in visually grounded dialogue based on autoregressive language modeling.
Results from our experiments on conversations from the visually grounded dialogue tasks \textsc{A Game Of Sorts} \citep[\textsc{AGOS},][]{willemsen_collecting_2022} and \textsc{PhotoBook} \citep[\textsc{PB},][]{haber_photobook_2019} were promising, showing that a text-only approach that involves the parameter-efficient fine-tuning of LLMs to generate annotated reproductions of utterances can be effective.
Moreover, we showed that providing the models with additional context from the dialogue history---that is, any messages that preceded the utterance under consideration---generally benefits performance.
Although these findings were largely consistent between the competing methods presented in this work, within our experimental setup the generative approach to information extraction using the fine-tuned, decoder-only \textsc{Llama} model was shown to consistently outperform the sequence labeling approach based on the fine-tuned, encoder-only \textsc{ModernBERT}.

Results from our cross-validation experiments showed that the models, on average, achieved better performance on the \textsc{PB-GOLD} than on the \textsc{AGOS} dataset.
The cross-dataset transfer experiments revealed a notable performance gap between the two datasets; fine-tuning on the \textsc{AGOS} data seemed to result in the models being better able to generalize beyond their specific conversational domain than when fine-tuning on the \textsc{PB-GOLD} data. 
These findings suggest that \textsc{AGOS} offers a more challenging testbed when it comes to MD, as it was explored in this work, than \textsc{PB-GOLD}. 
Given that the primary focus of the \textsc{PB} task is the correct identification of images, participants' language use is disproportionally reserved for referential purposes.
This was made apparent through a quantified characterization of the \textsc{PB-GOLD} mentions, indicating that mentions made up nearly two-thirds of the linguistic content of the dialogues.
In contrast, with image identification being a secondary objective, mentions make up just shy of one-third of the linguistic content of the \textsc{AGOS} dialogues.
In addition, mentions in the \textsc{PB-GOLD} dialogues are considerably longer, on average, than those in the \textsc{AGOS} dialogues.
When qualitatively examining the mentions from both datasets, it becomes clear that the incidence rate of mentions that resemble image caption-like descriptions is notably higher for \textsc{PB-GOLD} than for \textsc{AGOS}.
By and large, our findings suggest that \textsc{AGOS} offers its referring language use in a richer linguistic context than \textsc{PB-GOLD}, which aids the models' ability to generalize.

That being said, it would be reasonable to assume that the incidence rate of mentions in these task-oriented dialogues from both datasets is high compared to that of organic, non-task-oriented conversations. 
Conversations can go long stretches of time without the mention of a visually perceivable referent.
Our approach relies heavily on there being exploitable regularities in the linguistic context.
The extent to which conversations with comparatively sparse mention occurrences, and that take place outside of task-oriented settings, still exhibit such actionable patterns is, as of yet, unclear.
For both \textsc{AGOS} and \textsc{PB-GOLD}, the probability that a given linguistic expression (indirectly) points to a referent that is visually perceivable by at least one of the participants in the conversation is high, simply as a consequence of the situational context, as the images are the focal point of the conversations.
Discerning, from the linguistic context alone, whether an RE has such a referent becomes far more challenging, if not impossible, when the configuration of the visual context of the conversation is less constrained, more dynamic, and cannot be anticipated ahead of time.
In other words, we may still be able to extract mention candidates with a high degree of accuracy, but the number of false positives---by which we here mean any candidates that currently have no visually perceivable referents---is likely to be significantly higher; this outcome reminds of the high recall settings favored by aforementioned prior work on coreference resolution.

Inevitably, a general solution to the problem will require a cross-modal approach.
Although we make no assumptions regarding the manner of encoding, the visual information must somehow be incorporated to validate whether candidate mentions indeed have a referent in the visual context; even when the linguistic context strongly implies the existence of such a referent, we simply cannot be certain without a review of the visual context.  
Moreover, we are likely to see that end-to-end approaches will increasingly be favored over modular systems when it comes to addressing downstream tasks that have historically relied on some form of MD, but for which MD is simply a means to an end. %
Nevertheless, we expect that MD as a task in and of itself will remain relevant for niche applications for the foreseeable future. 
For one, it may continue to serve as a benchmark for the information extraction capabilities of models under varying conditions.
Perhaps more interestingly, however, are real-world applications, such as its use as an information extraction tool for corpus linguistics.

\section*{Limitations}
In this work, the focus has been on detecting REs that have a (visually perceivable) referent in the visual context of a conversation.
Only singletons and mentions in an identity relation were considered, contingent on their referent being one or more of the images in the visual context.
It is worth noting that there are some consequential differences between the images used by \textsc{AGOS} and \textsc{PB}.  
Where the focus of each \textsc{AGOS} image was on (an iconic view of) one entity from some image category, \textsc{PB} images depicted more complex scenes, purposely featuring multiple entities from different image categories.
Perhaps unsurprisingly, when the task involves identification within a visually grounded conversational context, we find that the more complex the scene, the more frequently we have to consider a bridging relationship between mentions as a surrogate for identity.
This highlights a complication with respect to the annotation of this domain that becomes increasingly problematic: the noisier (or more complex) the language use, the more ambiguous the boundaries.
We expect this to be even more evident in unrestricted, spoken dialogue.

Regarding our cross-validation experiments, results were based on a five-fold split of the datasets.
The \textsc{AGOS} dataset has a preferred partitioning that ensures minimal data leakage between the training and test data.
For \textsc{PB-GOLD}, however, we did not find a sensible, deterministic split, as even when image domains were seemingly mutually exclusive, in reality there were frequent intrusions from other image categories.
For instance, \textit{people}---which happens to be one of the author-defined image categories---are present in the vast majority of the photographs, often as salient entities, and frequently referenced as a result.
Although we do not believe this has affected our overall conclusions, the random splitting may have resulted in inflated scores in the \textsc{PB-GOLD} cross-validation experiments.
In addition, the language used in the dialogues from both datasets is exclusively English, meaning the experiments reported in this paper do not provide explicit insight into the extent to which the approach generalizes to other languages.

Finally, we have evaluated the proposed approach with one LLM undergoing a parameter-efficient fine-tuning regimen.
We have not investigated performance differences between full-parameter and parameter-efficient fine-tuning, nor have we tested the extent to which other generative LLMs are able to perform the task.
In addition, more exhaustive hyperparameter tuning has the potential to improve results further.
It is conceivable that more optimal hyperparameters exist that could narrow the observed performance gap between \textsc{Llama} and \textsc{ModernBERT} on this task.
However, it would mainly serve to underscore the general importance of the linguistic context and demonstrate the viability of either approach.

\section*{Acknowledgements}
This work was partially supported by the Wallenberg AI, Autonomous Systems and Software Program (WASP) funded by the Knut and Alice Wallenberg Foundation. 
The authors would like to thank Dmytro Kalpakchi, Jim O'Regan, Travis Wiltshire, Chris Emmery, Martina Rossi, and the anonymous reviewers for their helpful comments.

\bibliography{references}

\appendix

\section{Hyperparameters}
\label{sec:hyperparameters}

As a starting point for hyperparameter optimization, we took note of hyperparameters reported in prior work \citep[e.g.,][]{hu_lora_2022,dettmers_qlora_2023,warner_smarter_2024}, performing minimal tuning mostly within suggested ranges.

\begingroup
\renewcommand{\arraystretch}{1}
    \begin{table}[h!]
        \centering
        \footnotesize
        \begin{tabular}{l|c}\toprule
            Epochs & $2$ \\
            Batch size & $8$ \\
            Learning rate (LR) & $1$e-$4$ \\
            LR scheduler type & \texttt{cosine} \\
            Warmup ratio & $0.1$ \\
            \midrule
            LoRA $r$ & $16$ \\
            LoRA $\alpha$ & $16$ \\
            LoRA dropout & $0$ \\
            LoRA target modules & \texttt{*\_proj}, \texttt{lm\_head} \\
            \bottomrule
        \end{tabular}
        \caption{\label{table:hyperparameters-llama}
            Hyperparameters for QLoRA fine-tuning of \textsc{Llama 3.1 8B}. We use default values if not otherwise specified.
        }
    \end{table}
\endgroup

\begingroup
\renewcommand{\arraystretch}{1}
    \begin{table}[h!]
        \centering
        \footnotesize
        \begin{tabular}{l|c}\toprule
            Epochs & $4$ \\
            Batch size & $8$ \\
            Learning rate & $8$e-$5$ \\
            Gradient accumulation steps & $8$ \\
            Warmup ratio & $0.1$ \\
            Weight decay & $8$e-$6$ \\
            \bottomrule
        \end{tabular}
        \caption{\label{table:hyperparameters-modernbert}
            Hyperparameters for fine-tuning of \textsc{ModernBERT}-large. We use default values if not otherwise specified.
        }
    \end{table}
\endgroup

\section{Training example}
\label{sec:training-example}
The following is an example of a training sample from the \textsc{AGOS} dataset---for a context window of size $3$---that was used to fine-tune \textsc{Llama 3.1}:

\begin{quote}
\noindent{\color{purple}B: Clear, I think my second choice would be the light grey one, which looks like in old style.{\color{blue}\verb|\n|}A: I agree, its bottom seems to be pretty high as well.{\color{blue}\verb|\n|}B: yeap!{\color{blue}\verb|\n|}B: then, for the third one, is the dark grey one okay?{\color{blue}\verb|\n\n|}B: then, for the third one, is the dark grey one okay? {\color{orange}\verb|->|} B: then, for the third one, is {\color{orange}\verb|>>|} the dark grey {\color{orange}\verb|<<|} one okay?}    
\end{quote}

\noindent Messages in the linguistic context are separated by single newline characters ({\color{blue}\verb|\n|}).
Each message is prepended with a token indicating the speaker (either {\color{purple}A} or {\color{purple}B}).
The message we want annotated is separated from the linguistic context by two newline characters ({\color{blue}\verb|\n\n|}).
This message is followed by an inference token ({\color{orange}\verb|->|}). 
The inference token is then followed by the annotated message, with span boundary tokens indicating the start ({\color{orange}\verb|>>|}) and end ({\color{orange}\verb|<<|}) of the mention span.

\end{document}